%% file: main.tex
\documentclass{article}
\usepackage{spconf,amsmath,graphicx,amssymb}
\usepackage{multirow}
\usepackage{arydshln}

\usepackage{xcolor}
\usepackage{orcidlink}
\usepackage{booktabs}
\usepackage{makecell}
\usepackage{nicematrix}
\usepackage{subfig}
\usepackage{float}
\usepackage{algorithm,algorithmic}
\usepackage{colortbl}

\usepackage{caption}
\title{UMA-Split: Unimodal Aggregation for both English and Mandarin Non-Autoregressive Speech Recognition}
\name{Ying Fang$^{1, 2}$, Xiaofei Li$^{2,3}$\sthanks{Corresponding author.}}
\address{
$^{1}$ Zhejiang University, China \\
$^{2}$ School of Engineering, Westlake University, China \\
$^{3}$ Institute of Advanced Technology, Westlake Institute for Advanced Study, China}

\begin{document}
\fontsize{9.5pt}{\baselineskip}\selectfont
\maketitle
\begin{abstract}
This paper proposes a unimodal aggregation (UMA) based non-autoregressive model for both English and Mandarin speech recognition. The original UMA explicitly segments and aggregates acoustic frames (with unimodal weights that first monotonically increase and then decrease) of the same text token to learn better representations than regular connectionist temporal classification (CTC). However, it only works well in Mandarin. It struggles with other languages, such as English, for which a single syllable may be tokenized into multiple fine-grained tokens, or a token spans fewer than 3 acoustic frames and fails to form unimodal weights. To address this problem, we propose allowing each UMA-aggregated frame map to multiple tokens, via a simple split module that generates two tokens from each aggregated frame before computing the CTC loss. Experiments verify the proposed model’s effectiveness: it outperforms other advanced non-autoregressive models, and even matches the hybrid CTC/attention autoregressive model on LibriSpeech (English, 2.22\%/4.93\% WER for test clean/other with the 149 M model) and on AISHELL-1 (Mandarin, 4.43\% CER).
\end{abstract}

\input{Sections/1_introduction}

\input{Sections/2_methods}
\input{Sections/3_experiments}

\input{Sections/4_results}

\input{Sections/5_conclusion}

\vfill\pagebreak

\fontsize{9pt}{\baselineskip}\selectfont
\bibliographystyle{IEEEbib}
\bibliography{refs}

\end{document}

%% file: Sections/1_introduction.tex
\vspace{-1em}
\section{Introduction}
\label{sec:intro}

During the past decade, end-to-end (E2E) speech recognition models have made significant advancements \cite{prabhavalkar2023end}. A core challenge of E2E ASR models is aligning input acoustic feature frames with output text tokens, and mainstream models address this with distinct solutions. Attention-based encoder-decoder (AED) \cite{chan2016listen} models leverage cross-attention between inputs and outputs to determine which acoustic frames to focus on when generating the next token. Recurrent Neural Network Transducer (RNN-T) \cite{graves2012sequence} introduces blank tokens to facilitate alignment. It leverages a joint network for frame-by-frame token prediction, where predictions rely on both acoustic feature frames and outputs of the prediction network (conditioned on previously predicted tokens). Both models are autoregressive (AR) and typically use beam search decoding to enhance recognition accuracy, resulting in slow inference speeds.

In contrast, the connectionist temporal classification (CTC) \cite{graves2006connectionist} model is non-autoregressive (NAR). It predicts tokens in parallel, relying solely on acoustic feature frames. This frame-level independence enables faster inference, but at the cost of reduced recognition performance.
\cite{higuchi2021comparative} compares the performance of various NAR methods (Mask-CTC \cite{higuchi2020mask}, Intermediate CTC \cite{lee2021intermediate}, Self-conditioned CTC \cite{nozaki2021relaxing}, CIF \cite{dong2020cif}, etc.) on multiple English datasets up to 2021. Among them, Self-conditioned CTC performed best on most datasets: it embeds intermediate-layer CTC predictions into the forward flow to relax CTC’s independence assumption, while using the intermediate CTC loss. The study further notes that combining different NAR techniques boosts performance.

Over the past four years, several NAR methods \cite{Gao2022ParaformerFA, fang2024unimodal, zhuang2025effectiveasr} have been designed to explicitly aggregate acoustic information monotonically. They achieve performance comparable to AR baselines in Mandarin ASR. Among them, Paraformer \cite{Gao2022ParaformerFA} employs a Continuous Integrate-and-Fire (CIF) predictor to estimate the number of output tokens and integrate continuous acoustic feature frames. Paraformer-v2 \cite{an2024paraformer} observes that the CIF predictor fails to accurately estimate the number of English byte-pair encoding (BPE) tokens, so it replaces the CIF module with a CTC posterior module to enable English speech recognition.

This work proposes a unimodal aggregation (UMA) based model for NAR ASR on both English and Mandarin. As proposed in our prior work \cite{fang2024unimodal}, UMA explicitly segments and aggregates acoustic feature frames (with unimodal weights that first monotonically increase and then decrease) belonging to the same text token, thus learning better feature representations for text tokens compared to the regular CTC. However, it faces challenges when applied to languages other than Mandarin. 
In Mandarin ASR, tokens are Chinese characters—each corresponding to a complete long syllable, thus the continuous frames belonging to one syllable/token can be well aggregated. 
In other languages (e.g., English with BPE tokens), by contrast, a single syllable may be tokenized into multiple tokens, leaving CTC alignment struggling to learn which token each UMA-aggregated frame should map to. Additionally, fine-grained tokens may span fewer than 3 acoustic frames, failing to form unimodal weights.
To address these difficulties, this work proposes allowing each UMA-aggregated frame to map to multiple tokens, which is realized by designing a simple split module to generate two tokens from each UMA-aggregated frame before computing the CTC loss. 
Experiments on LibriSpeech (English) demonstrate that the proposed model can effectively conduct unimodal aggregation and then map one UMA-aggregated frame to multiple tokens. 
Overall, the proposed model achieves superior ASR performance for both English and Mandarin, and even reaches performance comparable to the hybrid CTC/AED AR model.

%% file: Sections/2_methods.tex
\begin{figure}[t]
\centering
\fontsize{9.5pt}{\baselineskip}\selectfont
\centerline{\includegraphics[width=0.7\columnwidth]{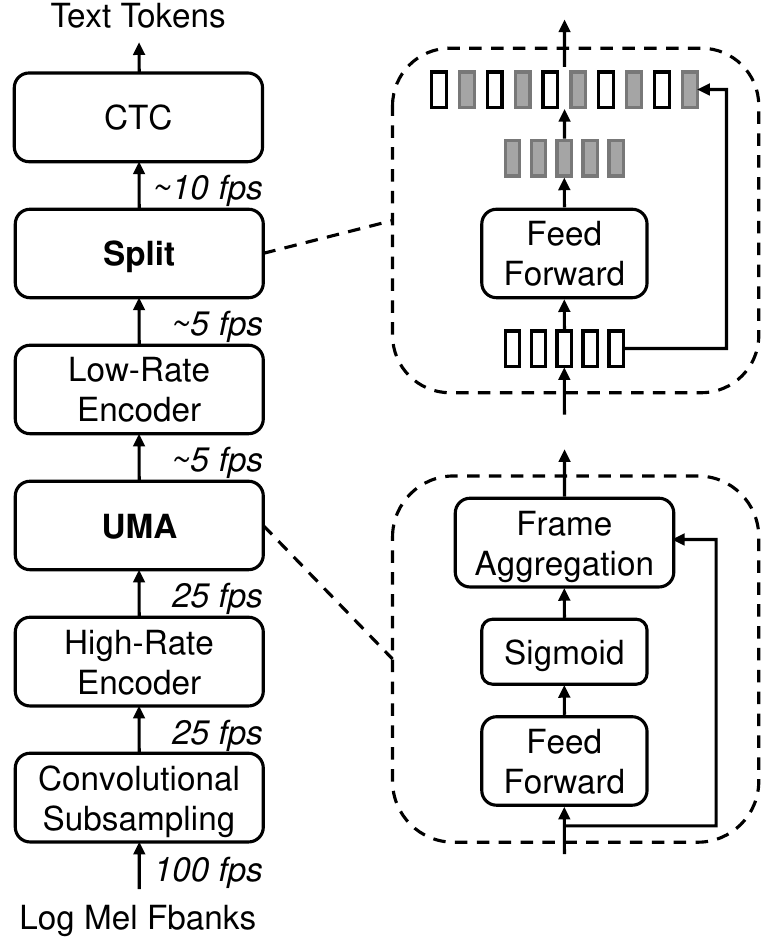}}
\caption{Model architecture. Frame rates (frames per second, fps) are computed when applying a 10 ms STFT frame-shift.}
\label{fig:model}
\vspace{-0.4cm}
\end{figure}

 \section{Method}
\label{sec:method}
The general architecture of the proposed model is shown in Fig.~\ref{fig:model}, consisting of six modules: convolutional subsampling, high-rate encoder, UMA module, low-rate encoder, split module, and CTC. The input of the model is log Mel filter-bank features, with a frame rate of 100 fps (frames per second) or 125 fps, depending on whether the STFT frame shift is 10 ms or 8 ms. A convolutional subsampling module is employed to downsample the feature sequence by a factor of 4, resulting in an output frame rate of 25 fps or 31.25 fps. 

\vspace{-1em}
\subsection{High-Rate Encoder}
The high-rate encoder generates acoustic features while preserving the frame rate after convolutional subsampling. It can be implemented with any sequence modeling network; for the experiments in this paper, we adopted the E-Branchformer encoder block \cite{kim2023branchformer}, given its superior performance in ASR tasks.

\vspace{-1em}
\subsection{Unimodal Aggregation Module}
In previous work \cite{fang2024unimodal, fang2025mamba}, UMA has been validated as effective for Mandarin speech recognition. In this work, we further improve its structure and extend its application to English.

UMA dynamically segments and aggregates the output feature sequence $\mathbf{e}_t^h, t=1, \dots, T$ of the high-rate encoder. For each time step, a scalar aggregation weight $\alpha_t$ is predicted using a feed-forward network (FFN) combined with a sigmoid activation, formulated as:
\begin{align}
\alpha_t=\text{Sigmoid}(\text{FFN}(\mathbf{e}_t^h)),
\end{align}
Consistent with the design in \cite{vaswani2017attention, gulati2020conformer}, the FFN comprises two linear layers interleaved with a Swish activation. The first linear layer applies a dimensional-expansion factor of 2, while the second projects to a 1-dimensional representation, which is followed by the sigmoid activation to generate the weight.

A timestep $t$ satisfying $\alpha_t\le \alpha_{t-1} \ \text{and} \ \alpha_t \le \alpha_{t+1}$ is defined as a UMA valley, with $t=0$ and $t=T$ also included.
The time index of the $i$-th UMA valley is denoted as $\tau_{i}\in[1, T]$, and we aggregate the feature frames between two consecutive UMA valleys using UMA weights as:
\begin{align}
\label{eq:uma}
\mathbf{c}_i=\frac{\sum_{t=\tau_{i}}^{\tau_{i+1}}\alpha_t \mathbf{e}_t^h}{\sum_{t=\tau_i}^{\tau_{i+1}}\alpha_t}.
\end{align}

The UMA valleys partition the feature frames into $i=1,\dots,I$ segments, so the length reduction rate is $T/I$. 
A UMA example is shown in Fig.~\ref{fig:uma}. This capability of dynamic segmentation and aggregation is automatically learned during training, without extra supervision regarding the length reduction rate.

\vspace{-1em}
\subsection{Low-Rate Encoder}
After the UMA, the sequence length is dynamically shortened. Influenced by factors including the text token rate, language type, and original frame rate, the post-UMA frame rate falls within the 4–7 fps range. The aggregated feature sequence is subsequently processed by a low-rate encoder, which preserves the frame rate. Our low-rate encoder consists of 6 Transformer encoder blocks \cite{vaswani2017attention}. The outputs are denoted as $\mathbf{e}_i^l \ (i=1, \dots, I)$.

\vspace{-1em}
\subsection{Split Module}
As mentioned in Section \ref{sec:intro}, to address the difficutlies of applying UMA to English, we propose allowing each UMA-aggregated frame to involve and generate two text tokens. This is realized by simply split one UMA-aggregated frame into two frames, with each possibly corresponding to one non-blank token.  
The output of the split module, denoted as $\mathbf{s}_j \ (j=1, \dots, 2I)$, is defined as: 
\begin{align}
\label{eq:split}
\mathbf{s}_j = 
\begin{cases} 
\text{LayerNorm}(\mathbf{e}_{i}^l) & \text{for } j=2i-1, \\
\text{LayerNorm}\left(\text{FFN}(\mathbf{e}_{i}^l)\right) & \text{for } j=2i.
\end{cases}
\end{align} 
The FFN employed here consists of two linear layers, with the first layer applying a dimensional-expansion factor of 4, and the second layer mapping back to the original dimension. 

\begin{figure}[t]
\centering
\fontsize{9.5pt}{\baselineskip}\selectfont
\centerline{\includegraphics[width=1.0\columnwidth]{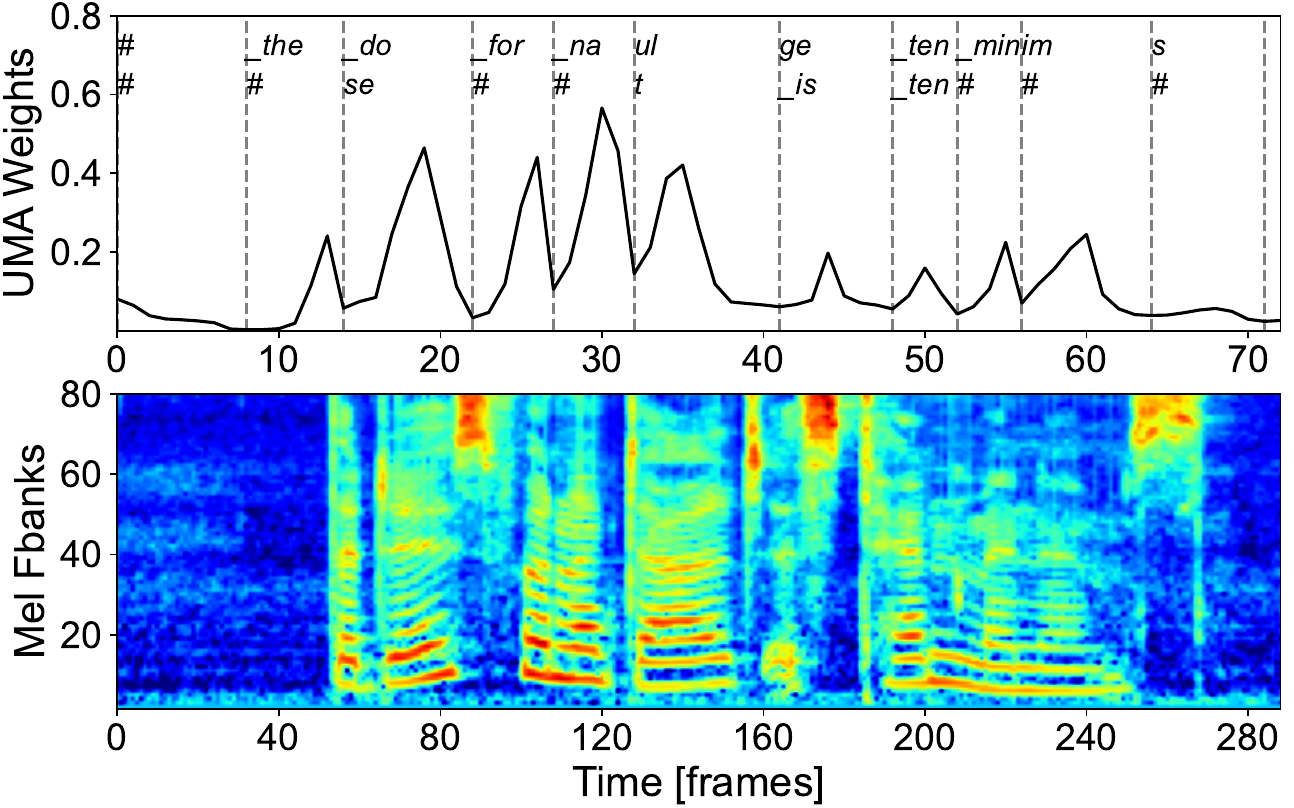}}
\caption{An English example of UMA weights for the BPE 5000 Base model. Ground truth text: \textit{the dose for an adult is ten minims}. \textit{\#} denotes blank token. 
}
\label{fig:uma}
\vspace{-0.4cm}
\end{figure}

We don't apply any explicit supervision to whether one UMA-aggregated frame would involve and generate two non-blank tokens; instead, this is learned automatically through the final CTC loss applied to the split sequence, i.e. $\mathbf{s}_j \ (j=1, \dots, 2I)$. In the example shown in Fig.~\ref{fig:uma}, three cases may occur for the two tokens after splitting: 1) Both are blank tokens; 2) Both are identical non-blank tokens, or one is a blank token—either case results in 1 non-blank token total (e.g., [\textit{\_ten,\_ten}], [\textit{\_the,\#}]); 3) The two are distinct non-blank tokens, leading to 2 non-blank tokens total (e.g., [\textit{\_do,se}]).

\vspace{-1em}
\subsection{Loss Function}
UMA's performance depends on the accuracy of estimating the span of acoustic feature frames of tokens, and integrating the Self-conditioned CTC (SC-CTC) method \cite{nozaki2021relaxing} can enhance the accuracy. 
Self-conditioned CTC adds CTC predictions from intermediate layers into the input of subsequent layers, aiming to condition the subsequent and final CTC predictions on these intermediate ones. We apply this method prior to the UMA module in the high-rate encoder, allowing the prediction of UMA weights to be conditioned on the intermediate CTC predictions. Specifically, we integrate conditioning layers into the mid-, three-quarter-, and final layers of the high-rate encoder. For the 2nd and 4th layers of the low-rate encoder, the intermediate CTC loss (no conditioning) is applied. Note that the intermediate hidden units here are fed into the split module before computing the intermediate CTC loss.  
The average of these five intermediate CTC losses gives $L_{\mathbf{inter}}$, combined with the CTC loss of the final output, it forms the overall training loss: $\mathcal{L} = 0.5({L_{\mathbf{CTC}}} + {L_{\mathbf{inter}}})$.

%% file: Sections/3_experiments.tex
\section{Experiments}
\label{sec:experiments}

\subsection{Dataset}
We evaluated the proposed method on two widely used datasets: 1) LibriSpeech \cite{panayotov2015librispeech}, an ~1,000 hours English audiobook speech dataset; 2) AISHELL-1 \cite{bu2017aishell}, an 178 hours Mandarin Chinese read speech dataset. For tokenization, Aishell-1 uses 4,233 Chinese characters, while LibriSpeech primarily uses 5,000 BPE tokens.  The BPE vocabulary is generated from the text of the training data. We additionally conducted experiments on LibriSpeech to explore the impact of BPE vocabulary size on UMA. 

\subsection{Experimental Setups}
ESPnet toolkit \cite{watanabe2018espnet} is used for all experiments. We released our codes on \footnote{https://github.com/Audio-WestlakeU/UMA-ASR}. The AR baseline is the hybrid CTC/AED model \cite{watanabe2017hybrid}, with all encoders being E-Branchformer \cite{kim2023branchformer}. We also noted that CTC networks trained via the hybrid CTC/AED method outperform those trained solely with CTC loss. Therefore, the baseline CTC models used for comparison are the models trained via the hybrid CTC/AED loss. Specifically, during inference, the AED head is not used, and only the CTC encoder is used. Given that the proposed model incorporates the SC-CTC mechanism, we also compared it with the original SC-CTC model \cite{nozaki2021relaxing} on LibriSpeech. Additionally, we directly quote results from several recent studies \cite{peng2022branchformer, kim2023branchformer, yaozipformer, an2024paraformer, zhuang2025effectiveasr, fang2024unimodal} on the same datasets to enable broader comparisons.

Note that at the early stage of training, unstable UMA aggregation may cause some batch samples to yield output lengths shorter than the target text token sequence, and thus incomputable CTC loss. Thus, we only adopt the batch samples with computable CTC loss for training, enabling normal gradient descent and gradual learning of reasonable aggregation ratios. In preliminary experiments, we tried character tokens on LibriSpeech. However, the token rate was too high (14.33 tps), failing to form effective UMA weights.

For each dataset, we ensure identical training configurations to the baseline models. All inferences are without language models. The CTC/AED baseline employs beam search (with 60 beams on LibriSpeech and 10 beams on AISHELL-1), while all CTC-related methods (including ours) use greedy search.

\noindent\textbf{Input:} 
All input data are sampled at 16kHz. The STFT frame shift is 10 ms for LibriSpeech, and 8 ms for AISHELL-1. The model input features are 80-dimentional log Mel filter-banks.

\noindent\textbf{High-rate encoder:} 
We use two sizes of E-Branchformer High-rate encoder on LibriSpeech: the hyperparameters of (dimension, feedforward dimension, layers, and attention heads) are set to Base (256, 1024, 13, 4) and Large (512, 1024, 18, 8). On AISHELL-1, the hyperparameters are (256, 1024, 17, 4). The numbers of layers have been adjusted to match the comparison baseline to have roughly equivalent model sizes. 

\noindent\textbf{Low-rate encoder:}  
6 Transformer encoder blocks, with dimension and attention heads matching the respective high-rate encoder. All models use 2048 feedforward dimension.

\noindent\textbf{Optimizer:} 
We use the AdamW optimizer and warmup scheduler. The batch size, learning rate, and warmup steps follow ESPnet's recipes. For weight decay, we use 1e-6 for LibriSpeech and 1e-2 for AISHELL-1. The proposed model and the baselines are trained for the same number of steps. The model with averaged weights of the 10 best checkpoints is taken as the final model.

%% file: Sections/4_results.tex

\begin{table*}[t]
\centering
\vspace{-0.1cm}
\fontsize{9pt}{\baselineskip}\selectfont
\caption{UMA-Split model-related statistics. 
\textbf{Token rate}: Tokens per second (tps) calculated on test sets. 
\textbf{Non-blank}: Proportion of frames that output $\geq 1$ non-blank tokens after splitting (relative to all frames). 
\textbf{2-non-blank}: Proportion of frames that output 2 distinct non-blank tokens (relative to \textbf{Non-blank} frames). 
SC represents SC-CTC.}

\label{tab:uma_stat}
\setlength{\tabcolsep}{1mm}{
\begin{tabular}{c|lc|cc|cc|cc}
    \Xhline{1pt}
    \multirow{2}{*}{\textbf{Dataset}} 
    &{\textbf{Token type \&}} & \multirow{2}{*}{\textbf{Token rate}} & \multicolumn{2}{c|}{\textbf{Frame rate}}  & \multicolumn{2}{c|}{\textbf{Split ratio}} & \multirow{2}{*}{\textbf{Params}} & \textbf{Test (\%)}\\ 
    & {\textbf{vocabulary size}} &  & \textbf{before UMA} & \textbf{after UMA} & \textbf{non-blank} & \textbf{2-non-blank} &  & \textbf{CER or WER}\\ 
    \hline
    \multirow{1}{*}{AISHELL-1} 
     & Char 4233 & 2.90 tps & 31.25 fps & 5.91 fps & 49.4\% & 0\% & 46M & 4.43\\
    \hline
    \multirow{5}{*}{LibriSpeech}
    & BPE 500 & 5.37 tps & 25 fps &  6.16 fps & 73.2\% & 30.1\% & 39M & 2.75 / 6.45 \\
    & BPE 5000 (B) & 3.39 tps & 25 fps & 4.58 fps & 70.5\% & 8.3\% & 41M & 2.50 / 5.77\\
    & BPE 10000 & 3.11 tps & 25 fps & 4.38 fps &  68.7\% & 4.9\% & 43M & 2.49 / 5.73\\
    \cdashline{2-9}
    & BPE 5000 (w/o SC) & 3.39 tps & 25 fps & 4.98 fps & 61.5\% &  12.6\% & 39M & 2.90 / 6.53 \\
    & BPE 5000 (L) & 3.39 tps & 25 fps & 5.78 fps & 56.1\% & 7.6\% & 149M & 2.22 / 4.93\\
    \Xhline{1pt}
\end{tabular}
}
\vspace{-0.2cm}
\end{table*}

\begin{table}[t]
\vspace{-0.2cm}
\centering
\fontsize{9pt}{\baselineskip}\selectfont
\caption{Word error rate (WER,\%) on LibriSpeech. All without LM. The decoding beam size or AR models is 60. Test set results are divided into test\_clean / test\_other.}
\label{tab:librispeech}
\setlength{\tabcolsep}{1.5mm}{
\begin{tabular}{l|c|cc}
    \Xhline{1pt}
    \multicolumn{1}{c|}{\textbf{Model}} & {\textbf{Type}} &{\textbf{clean / other}} & \textbf{Params} \\ 
    \hline 
    E-Branchformer (B), hybrid \cite{kim2023branchformer} & AR & \textbf{2.49} / \textbf{5.61}  & 41M \\
    \hspace{0.8cm} CTC Infer w/o AED head & NAR & 3.20 / 7.09  & 29M\\
    Zipformer-M, CTC \cite{yaozipformer} & NAR & 2.52 / 6.02 & 64M \\
    Paraformer-v2 (S) \cite{an2024paraformer} & NAR & 3.4 / 8.0  & 50M \\
    E-Branchformer, SC-CTC & NAR & 2.62  / 6.16 & 43M \\
    UMA-Split (B) (prop.) & NAR & \textbf{2.50} / \textbf{5.77} & 41M \\ 
    \cdashline{1-4}
    E-Branchformer (L), hybrid \cite{kim2023branchformer} & AR & \textbf{2.14}  / \textbf{4.55}  & 149M \\ 
    \hspace{0.8cm} CTC Infer w/o AED head & NAR & 2.59 / 5.45  & 119M\\
    Zipformer-L, CTC \cite{yaozipformer} & NAR & 2.50 / 5.72 & 147M \\
    Paraformer-v2 (L) \cite{an2024paraformer} & NAR & 3.0 / 6.9  & 120M\\
    UMA-Split (L) (prop.) & NAR & \textbf{2.22} / \textbf{4.93} & 149M \\
    \Xhline{1pt}
\end{tabular}
}
\vspace{-0.2cm}
\end{table}

\begin{table}[h]
\vspace{-0.2cm}
\centering
\fontsize{9pt}{\baselineskip}\selectfont
\caption{Character error rate (CER,\%) on AISHELL-1. All without LM. AR model decoding beam size 10.}
\label{tab:aishell-1}
\setlength{\tabcolsep}{1.5mm}{
\begin{tabular}{l|c|ccc}
    \Xhline{1pt}
    \multicolumn{1}{c|}{\textbf{Model}} & {\textbf{Type}} & \textbf{dev} & \textbf{test} & \textbf{Params} \\ 
    \hline
    Branchformer (B), hybrid \cite{peng2022branchformer} & AR & 4.19 & \textbf{4.43}  & 45M\\
    E-Branchformer, hybrid & AR & \textbf{4.13}  & 4.53  & 57M \\ 
    \hspace{0.8cm} CTC Infer w/o AED head & NAR & 4.39 & 4.91  & 46M\\
    Paraformer-v2 (S) \cite{an2024paraformer} & NAR & 4.5 & 4.9  & 50M \\
    Zipformer-M, CTC\cite{yaozipformer} & NAR & 4.47 & 4.80 & 66M \\
    EffectiveASR Large \cite{zhuang2025effectiveasr} & NAR & 4.26 & 4.62  & 76M\\
    Original UMA Conformer \cite{fang2024unimodal} & NAR & 4.4 & 4.7  & 45M \\
    UMA-Split (prop.) & NAR & \textbf{4.15} & \textbf{4.43} & 46M \\ 
    \hspace{0.8cm} - w/o split module & NAR & 4.28 & 4.53 & 45M \\ 
    \Xhline{1pt}
\end{tabular}
}
\vspace{-0.3cm}
\end{table}

\subsection{Main Results}
\label{sec:result}
\noindent\textbf{LibriSpeech:} 
Table \ref{tab:librispeech} shows the results on LibriSpeech. Compared with other advanced NAR models, the proposed model achieves superior performance in both Base (41 M parameters) and Large (149 M parameters) configurations, which demonstrates that the proposed explicit frame aggregation with UMA weights indeed helps learn better token representations. Compared to the AR hybrid CTC/AED baseline, the proposed model achieves comparable performance across different parameter scales, with a 10× inference speedup \cite{fang2024unimodal}.

\noindent\textbf{AISHELL-1:} 
Table \ref{tab:aishell-1} shows the results on AISHELL-1. Consistent to the results presented in \cite{fang2024unimodal}, UMA outperforms other NAR models. 
Compared to the model without the split module, adding it brings a slight performance boost in Mandarin ASR. The superior results compared to other models confirm that the proposed model retains the original UMA's strong capability for Mandarin ASR.

\subsection{UMA-Split Analysis}
\label{sec:uma_analysis}
Table \ref{tab:uma_stat} presents UMA-Split model statistics across different configurations. 
On AISHELL-1, each token corresponds to a complete syllable. Even with the split module, UMA-aggregated frames still belong to a single token, so the 2-non-blank split ratio is 0\%.

On LibriSpeech, we compare different BPE vocabulary sizes with Base model. We find that as the vocabulary size increases, both the token rate and the frame rate after UMA decrease. A larger BPE vocabulary size has coarser-grained tokenization and larger average number of acoustic frames per token, which is easier for UMA learning.  
This results in a lower 2-non-blank split ratio. Although the split module allows one UMA-aggregated frame to map to 2 non-blank tokens, such mixed representation of 2 tokens is inferior than the representation of 1 token. Therefore, a lower 2-non-blank split ratio corresponds to a lower WER. 

Additionally, compared with the model without SC-CTC (w/o SC) on LibriSpeech, we can find that SC-CTC is helpful for better recognizing the span of acoustic feature frames of tokens and thus for reducing the 2-non-blank split ratio and WER. 
We also note that compared to the Base model, the Large model exhibits a higher frame rate after UMA and generates more blank tokens (with a lower non-blank split ratio), while the reasons for this phenomenon remains unclear to us.


%% file: Sections/5_conclusion.tex
\section{Conclusions}
\label{sec:conclusion}
We propose a UMA-based NAR ASR model for English and Mandarin. The original UMA faces challenges in languages other than Mandarin (e.g., English with BPE tokens), where fine-grained tokens may span fewer than 3 acoustic frames, failing to form unimodal weights. We address this by allowing each UMA-aggregated frame to map to two tokens via a simple split module. Experiments on LibriSpeech and AISHELL-1 show the proposed model achieves performance even comparable to that of the hybrid CTC/AED AR model.

%% file: refs.bib
@inproceedings{bu2017aishell,
  title={{AISHELL-1}: An open-source mandarin speech corpus and a speech recognition baseline},
  author={Bu, Hui and Du, Jiayu and Na, Xingyu and Wu, Bengu and Zheng, Hao},
  booktitle={O-COCOSDA},
  pages={1--5},
  year={2017}
}

@inproceedings{watanabe2018espnet,
  author={Shinji Watanabe and Takaaki Hori and Shigeki Karita and Tomoki Hayashi and Jiro Nishitoba and Yuya Unno and Nelson {Enrique Yalta Soplin} and Jahn Heymann and Matthew Wiesner and Nanxin Chen and Adithya Renduchintala and Tsubasa Ochiai},
  title={{ESPnet}: End-to-End Speech Processing Toolkit},
  year={2018},
  booktitle={Interspeech},
  pages={2207--2211},  
}

@inproceedings{graves2006connectionist,
  title={Connectionist temporal classification: labelling unsegmented sequence data with recurrent neural networks},
  author={Graves, Alex and Fern{\'a}ndez, Santiago and Gomez, Faustino and Schmidhuber, J{\"u}rgen},
  booktitle={ICML},
  pages={369--376},
  year={2006}
}

@inproceedings{gulati2020conformer,
  author={Anmol Gulati and James Qin and Chung-Cheng Chiu and Niki Parmar and Yu Zhang and Jiahui Yu and Wei Han and Shibo Wang and Zhengdong Zhang and Yonghui Wu and Ruoming Pang},
  title={{Conformer: Convolution-augmented Transformer for Speech Recognition}},
  year=2020,
  booktitle={Interspeech},
  pages={5036--5040},
  doi={10.21437/Interspeech.2020-3015}
}

@inproceedings{fang2024unimodal,
  title={Unimodal aggregation for {CTC}-based speech recognition},
  author={Fang, Ying and Li, Xiaofei},
  booktitle={ICASSP},
  pages={10591--10595},
  year={2024}
}

@inproceedings{dong2020cif,
  title={{CIF}: Continuous integrate-and-fire for end-to-end speech recognition},
  author={Dong, Linhao and Xu, Bo},
  booktitle={ICASSP},
  pages={6079--6083},
  year={2020}
}

@inproceedings{panayotov2015librispeech,
  title={Librispeech: an {ASR} corpus based on public domain audio books},
  author={Panayotov, Vassil and Chen, Guoguo and Povey, Daniel and Khudanpur, Sanjeev},
  booktitle={ICASSP},
  pages={5206--5210},
  year={2015},
}

@inproceedings{kim2023branchformer,
  title={{E-Branchformer}: Branchformer with enhanced merging for speech recognition},
  author={Kim, Kwangyoun and Wu, Felix and Peng, Yifan and Pan, Jing and Sridhar, Prashant and Han, Kyu J and Watanabe, Shinji},
  booktitle={SLT},
  pages={84--91},
  year={2023}
}

@article{vaswani2017attention,
  title={Attention is all you need},
  author={Vaswani, Ashish and Shazeer, Noam and Parmar, Niki and Uszkoreit, Jakob and Jones, Llion and Gomez, Aidan N and Kaiser, {\L}ukasz and Polosukhin, Illia},
  journal={Advances in neural information processing systems},
  volume={30},
  year={2017}
}

@inproceedings{nozaki2021relaxing,
  title={Relaxing the conditional independence assumption of {CTC}-based {ASR} by conditioning on intermediate predictions},
  author={Nozaki, Jumon and Komatsu, Tatsuya},
  booktitle={Interspeech},
  year={2021}
}

@article{watanabe2017hybrid,
  title={Hybrid {CTC/attention} architecture for end-to-end speech recognition},
  author={Watanabe, Shinji and Hori, Takaaki and Kim, Suyoun and Hershey, John R and Hayashi, Tomoki},
  journal={IEEE Journal of Selected Topics in Signal Processing},
  volume={11},
  number={8},
  pages={1240--1253},
  year={2017}
}

@inproceedings{yaozipformer,
  title={Zipformer: A faster and better encoder for automatic speech recognition},
  author={Yao, Zengwei and Guo, Liyong and Yang, Xiaoyu and Kang, Wei and Kuang, Fangjun and Yang, Yifan and Jin, Zengrui and Lin, Long and Povey, Daniel},
  booktitle={ICLR},
  year={2024}
}

@article{an2024paraformer,
  title={Paraformer-v2: An improved non-autoregressive transformer for noise-robust speech recognition},
  author={An, Keyu and Li, Zerui and Gao, Zhifu and Zhang, Shiliang},
  journal={Nat. Conf. Man- Mach. Speech Commun.}, 
  pages={1240--1253},
  year={2024}
}

@inproceedings{peng2022branchformer,
  title={Branchformer: Parallel {MLP}-attention architectures to capture local and global context for speech recognition and understanding},
  author={Peng, Yifan and Dalmia, Siddharth and Lane, Ian and Watanabe, Shinji},
  booktitle={ICML},
  pages={17627--17643},
  year={2022}
}

@inproceedings{zhuang2025effectiveasr,
  title={EffectiveASR: A Single-Step Non-Autoregressive Mandarin Speech Recognition Architecture with High Accuracy and Inference Speed},
  author={Zhuang, Ziyang and Miao, Chenfeng and Zou, Kun and Fang, Ming and Wei, Tao and Li, Zijian and Cheng, Ning and Hu, Wei and Wang, Shaojun and Xiao, Jing},
  booktitle={ICASSP},
  pages={1--5},
  year={2025}
}

@inproceedings{chan2016listen,
  title={Listen, attend and spell: A neural network for large vocabulary conversational speech recognition},
  author={Chan, William and Jaitly, Navdeep and Le, Quoc and Vinyals, Oriol},
  booktitle={ICASSP},
  pages={4960--4964},
  year={2016}
}

@article{graves2012sequence,
  title={Sequence transduction with recurrent neural networks},
  author={Graves, Alex},
  journal={Proc. Int. Conf. Mach. Learn.},
  year={2012}
}

@article{prabhavalkar2023end,
  title={End-to-end speech recognition: A survey},
  author={Prabhavalkar, Rohit and Hori, Takaaki and Sainath, Tara N and Schl{\"u}ter, Ralf and Watanabe, Shinji},
  journal={IEEE/ACM Transactions on Audio, Speech, and Language Processing},
  pages={325--351},
  year={2023},
}

@inproceedings{higuchi2021comparative,
  title={A comparative study on non-autoregressive modelings for speech-to-text generation},
  author={Higuchi, Yosuke and Chen, Nanxin and Fujita, Yuya and Inaguma, Hirofumi and Komatsu, Tatsuya and Lee, Jaesong and Nozaki, Jumon and Wang, Tianzi and Watanabe, Shinji},
  booktitle={ASRU},
  pages={47--54},
  year={2021}
}

@inproceedings{higuchi2020mask,
  title={Mask {CTC}: Non-autoregressive end-to-end ASR with {CTC} and mask predict},
  author={Higuchi, Yosuke and Watanabe, Shinji and Chen, Nanxin and Ogawa, Tetsuji and Kobayashi, Tetsunori},
  booktitle={Interspeech},
  year={2020}
}

@inproceedings{lee2021intermediate,
  title={Intermediate loss regularization for {CTC}-based speech recognition},
  author={Lee, Jaesong and Watanabe, Shinji},
  booktitle={ICASSP},
  pages={6224--6228},
  year={2021}
}

@inproceedings{Gao2022ParaformerFA,
  title={Paraformer: Fast and Accurate Parallel Transformer for Non-autoregressive End-to-End Speech Recognition},
  author={Zhifu Gao and Shiliang Zhang and Ian Mcloughlin and Zhijie Yan},
  booktitle={Interspeech},
  year={2022},
  url={https://api.semanticscholar.org/CorpusID:249712411}
}

@inproceedings{fang2025mamba,
  title={Mamba for streaming asr combined with unimodal aggregation},
  author={Fang, Ying and Li, Xiaofei},
  booktitle={ICASSP},
  pages={1--5},
  year={2025}
}
